\documentclass[letterpaper]{article} 
\usepackage{aaai2026}  
\newcounter{eqfnq}
\newcommand{\equalcontribq}{%
    \ifnum\value{eqfnq}=0%
        \thanks{These authors contributed equally.}%
        \setcounter{eqfnq}{\value{footnote}}%
    \else%
        \footnotemark[\value{eqfnq}]%
    \fi%
}
\usepackage{times}  
\usepackage{helvet}  
\usepackage{courier}  
\usepackage[hyphens]{url}  
\usepackage{graphicx} 
\usepackage{natbib}  
\usepackage{caption} 
\usepackage{algorithm}
\usepackage{algorithmic}
\usepackage{amsmath}
\usepackage{amssymb}
\usepackage{amsfonts}
\usepackage{booktabs}
\usepackage{bbding}  
\usepackage{pifont}  
\usepackage{multirow}
\usepackage{makecell} 
\usepackage{adjustbox}
\usepackage{subfig}

\usepackage{multirow}
\usepackage{booktabs}
\usepackage{amsmath}  
\usepackage{bbding}  
\usepackage{pifont}
\usepackage{algorithm}
\usepackage{algorithmic}
\usepackage{tikz}
\usepackage{adjustbox}
\usepackage{array}
\usepackage{amsthm}  
\newtheoremstyle{tightstyle}
  {1pt} 
  {1pt} 
  {\itshape} 
  {} 
  {\bfseries} 
  {.} 
  { } 
  {}
\theoremstyle{tightstyle}
\newtheorem{theorem}{Theorem}

\newcolumntype{C}[1]{>{\centering\arraybackslash}p{#1}}
\newcommand{\Comment}[1]{\hfill$\triangleright$ #1}

\usepackage{newfloat}
\usepackage{listings}
\DeclareCaptionStyle{ruled}{labelfont=normalfont,labelsep=colon,strut=off} 
\floatstyle{ruled}
\newfloat{listing}{tb}{lst}{}
\floatname{listing}{Listing}
\title{ReAlign: Text-to-Motion Generation via Step-Aware Reward-Guided Alignment}

\usepackage{bibentry}
\author{
    Wanjiang~Weng\textsuperscript{1,2}\equalcontribq, 
    Xiaofeng~Tan\textsuperscript{1,2}\equalcontribq, 
    Junbo~Wang\textsuperscript{3},
    Guo-Sen~Xie\textsuperscript{4},
    Pan~Zhou\textsuperscript{5},
    Hongsong~Wang\textsuperscript{1,2}\thanks{Corresponding Author.},
}
\affiliations{
    \textsuperscript{1}School of Computer Science and Engineering, Southeast University, Nanjing, China \\
    \textsuperscript{2}Key Laboratory of New Generation Artificial Intelligence Technology and Its Interdisciplinary Applications (Southeast University), Ministry of Education, China \\
    \textsuperscript{3}School of Software, Northwestern Polytechnical University, Xi'an, China \\
    \textsuperscript{4}School of Computer Science and Engineering, Nanjing University of Science and Technology, Nanjing, China \\
    \textsuperscript{5}Singapore Management University, Singapore \\
    \{wjweng, xiaofengtan, hongsongwang\}@seu.edu.cn;jbwang@nwpu.edu.cn;gsxiehm@gmail.com;panzhou3@gmail.com}

\begin{document}
\maketitle

\begin{abstract}
Text-to-motion generation, which synthesizes 3D human motions from text inputs, holds immense potential for applications in gaming, film, and robotics. Recently, diffusion-based methods have been shown to generate more diversity and realistic motion. However, there exists a misalignment between text and motion distributions in diffusion models, which leads to semantically inconsistent or low-quality motions. To address this limitation, we propose \textbf{Re}ward-guided sampling \textbf{Align}ment ({\textbf{ReAlign}}), comprising a step-aware reward model to assess alignment quality during the denoising sampling and a reward-guided strategy that directs the diffusion process toward an optimally aligned distribution. This reward model integrates step-aware tokens and combines a text-aligned module for semantic consistency and a motion-aligned module for realism, refining noisy motions at each timestep to balance probability density and alignment. Extensive experiments of both motion generation and retrieval tasks demonstrate that our approach significantly improves text-motion alignment and motion quality compared to existing state-of-the-art methods.
\end{abstract}

\begin{links}
    \link{Code}{https://wengwanjiang.github.io/ReAlign-page}
\end{links}

\section{Introduction}\label{intro} 
With the rising demand for realistic 3D human motions in gaming, filmmaking, virtual reality, and robotics, along with recent advances in motion modeling~\cite{motiongpt,wang2025foundation}, there is an increasing need for intuitive and controllable motion generation techniques.
Text-to-motion generation, which aims to synthesize human motion directly from natural language descriptions, has emerged as a key research topic~\cite{Chen2023, motionlcm, Guo2022, motiongpt, wu2025finemotion, wu2025mg}.

Diffusion has emerged as the mainstream approach for text-driven motion generation~\cite{Tevet2023,Zhang2024}. However, diffusion-based models often struggle with text-motion alignment due to their reliance on text embeddings encoded by CLIP~\cite{radford2021learning}, which is trained on text-image pairs rather than text-motion pairs. Consequently, these models often fail to capture the semantic alignment between text and motion. The motions synthesized by most existing diffusion-based methods lack coherence with the input descriptions (see Figure~\ref{fig:intro}).

Prior works aiming to improve text-to-motion alignment, such as reinforcement learning with reward functions~\cite{han2024reindiffuse, liu2024motionrl, tan2025sopo}, primarily focus on fine-tuning generative models to enhance motion quality. These approaches lack the capability to handle noisy motion inputs. Moreover, the misalignment issue should be addressed during the denoising process itself, rather than corrected retrospectively after the final motion is generated.

Another limitation of existing diffusion-based methods is that, although they can generate high-quality motions, the generated motions may still lack smoothness and realism in some cases. During the reverse diffusion process, motion generation relies solely on the diffusion model without access to real motion references for guidance. To address this issue, we shift our focus to motion retrieval. In fact, motion generation and retrieval are closely related tasks; however, most existing works investigate them separately, with few efforts dedicated to exploring their interconnection or developing a unified model for both tasks.

To address these problems, we propose a novel \textbf{Re}ward-guided sampling \textbf{Align}ment strategy (\textbf{ReAlign}) to enhance text-motion alignment quality with the guidance of a well-aligned reward distribution. We derive the reward distribution from a step-aware reward comprising two modules: a text-aligned module to ensure semantic consistency, and a motion-aligned module to assess realism. Together, these modules adapt to noisy motions and variations across timesteps, guiding diffusion model toward a distribution that not only maximizes probability density but also maintains strong text-motion alignment. By explicitly addressing both semantic misalignment and motion quality degradation, this approach improves the coherence and realism of the generated motion. 
The proposed reward model is plug-and-play and can be seamlessly integrated into any motion diffusion model without requiring additional fine-tuning. Our main contributions are as follows:
\begin{itemize}
\item[$\bullet$] \textbf{Theoretical reward-guided denoising analysis:} We theoretically demonstrate that the reward gradient, derived from both text-aligned and motion-aligned rewards, progressively influences the denoising process, guiding the sampling trajectory toward a distribution that better reflects the intended motion semantics.
\item[$\bullet$] \textbf{Versatile module for diffusion-based generation:} We propose ReAlign, which comprises a step-aware reward model and a reward-guided sampling strategy to improve text-motion alignment. Extensive experiments demonstrate that our approach significantly enhances existing diffusion-based motion generation models.
\end{itemize}

\section{Related Works}\label{rw}
\noindent\textbf{Alignment in Text-to-Motion Generation.} Text-to-motion generation represents a critical task in computer vision, exhibiting rapid advancements in recent years~\cite{zhang2025large, zhang2023finemogen, zhang2023remodiffuse,yuan2025mogents}. Diffusion models have been adopted for text-driven motion generation \cite{Tevet2023,Zhang2024}. MotionLCM~\cite{motionlcm} refines motion-latent diffusion to enable precise spatiotemporal control via few-step inference.

Alignment represents a versatile technique widely employed across the domains of language modeling \cite{rafailov2023direct}, image generation \cite{wallace2024diffusion}, and policy optimization~\cite{chen2023score}. Recently, human preference alignment is studied in text-to-motion generation. ReinDiffuse~\cite{han2024reindiffuse} refines the diffusion model through reinforcement learning to enhance the physical plausibility of generated motions. MotionRL~\cite{liu2024motionrl} focus aligning human preferences using the proposed multi-reward reinforcement learning framework. SoPo~\cite{tan2025sopo} combines the strengths of online and offline direct preference optimization to overcome their individual shortcomings, delivering enhanced motion generation quality and preference alignment.
However, these methods focus on fine-tuning generative models to align preferences or enhance motion quality without explicitly addressing text-motion misalignment. In contrast, we tackle this issue with a plug-and-play reward model in the inference process.

\noindent \textbf{Diffusion-Based Reward-Guided Generation.} Guidance in diffusion models can be driven by either derivative-free or gradient-based reward models. Derivative-free methods include Sequential Monte Carlo (SMC)-based guidance~\cite{dou2024diffusion,cardoso2024monte} and Value-Based Importance Sampling~\cite{li2025towards}, where the former samples across the entire batch, while the latter performs sampling independently for each sample without global interaction. The typical gradient-based reward is classifier guidance~\cite{classg}.
Inference-time guidance is specifically designed for discrete diffusion models~\cite{uehara2025reward}. Exact sampling from the optimal policy is feasible within discrete diffusion models under certain limited scenarios. A practical implementation of classifier guidance for discrete diffusion models proposed~\cite{nisonoffunlocking}. Search-based algorithms~\cite{wan2024alphazero} are also proposed to enhance inference alignment. For image synthesis, Liu et al.~\shortcite{liu2023more} introduce a unified framework for multi-modal guidance with both language and image input. While reward-guided generation has received little attention in motion synthesis, this work addresses that gap.

\section{Methods}\label{methods}
\label{sec:reward_guided_alignment}

\subsection{Motivation and Overview Framework}
\label{sec:431}

 \begin{figure*}[tbp]
   \centering
   \includegraphics[width=0.9\textwidth]{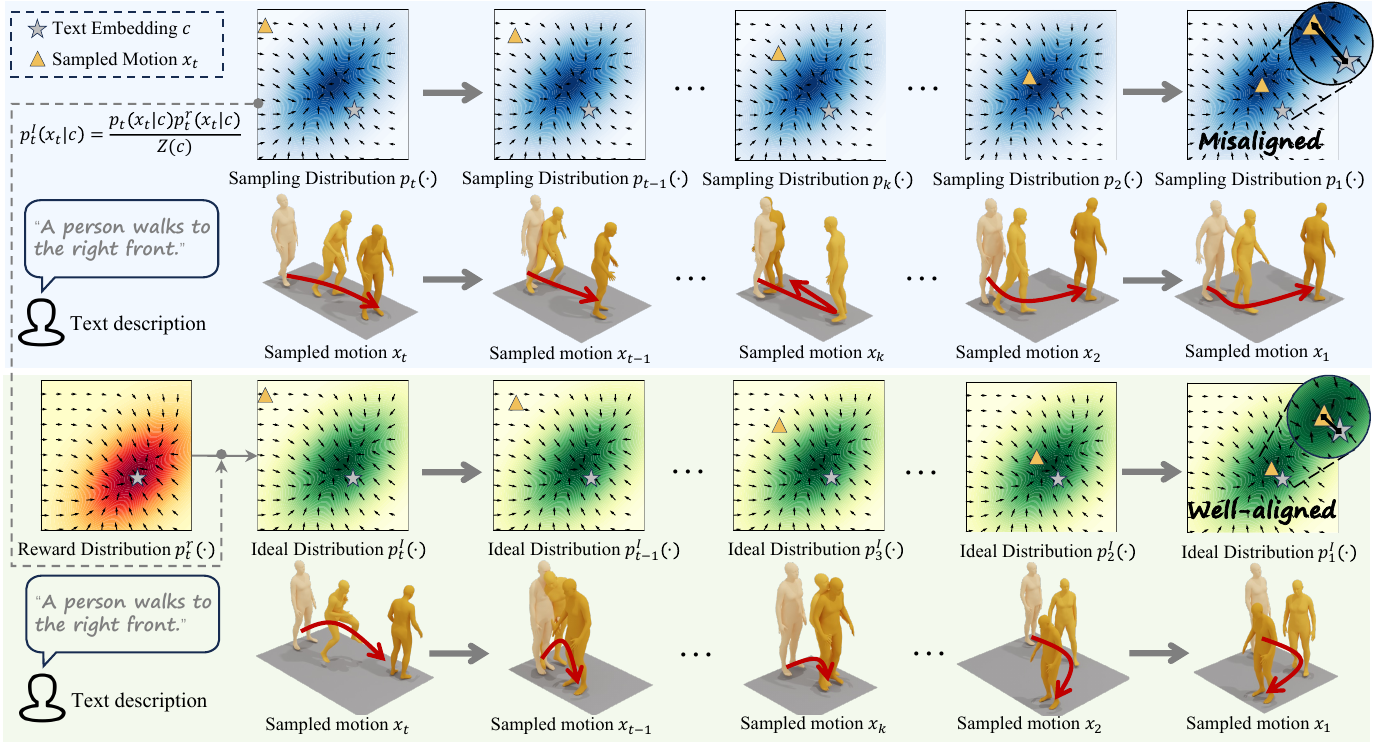}
   \caption{Illustration of the sampling process in diffusion-based motion generation frameworks. The blue region represents the sampling distribution $p_t(\cdot)$ learned by the diffusion model, while the green region depicts the ideal sampling distribution $p_t^I(\cdot)$ achieved by incorporating our proposed reward-guided sampling strategy with the sampling distribution $p_t(\cdot)$.}
   \label{fig:sampling}
\end{figure*}

\noindent\textbf{Preliminaries.} Existing diffusion-based motion generation methods \cite{Chen2023, Tevet2023} operate via a forward process and a reverse process. The forward process gradually adds noise into the real motion distribution $ p_\text{data}(\cdot) $ over timestep, and can be modeled as a stochastic differential equation (SDE) \cite{songscore}:  
\begin{equation}
	\mathrm{d}\mathbf{x} = \mathbf{f}(\mathbf{x},t) \mathrm{d}t + g(t) \mathrm{d}\mathbf{w},
\end{equation}  
where $ t $ is  timestep, $ \mathbf{f}(\cdot,\cdot) $ and $ g(\cdot) $ are the drift and diffusion coefficients, and $ \mathbf{w} $ is the Wiener process. For reverse process, motions $ \mathbf{x} $ are generated via trajectory sampling~\cite{songscore}:  
\begin{equation}
	\label{eq:ideal_reverse}
	\mathrm{d}\mathbf{x} = [\mathbf{f}(\mathbf{x},t)- g (t)^2\nabla \log p_t(\mathbf{x})] \mathrm{d}t + g(t) \mathrm{d}\mathbf{w},
\end{equation}  
where $ \nabla \log p_t(\mathbf{x}) $ is the score function of $ p_t(\mathbf{x}) $, directing sampling toward higher-density regions.

\noindent\textbf{Motivation.} While existing text-to-motion diffusion models enable motion generation with high-quality, they often fail to generate motions that accurately align with textual descriptions. For example, as illustrated in Figure~\ref{fig:sampling}, the diffusion model prompted to generate a person walking forward to the right may instead veer left. This misalignment arises as the sampling distribution $ p_t(\mathbf{x}) $, learned from the diffusion, prioritizes high-probability regions over semantic fidelity.

Upon analyzing the diffusion sampling process (Figure~\ref{fig:sampling}), we identify a key issue: sampled motions $ \mathbf{x}_t $ (stars) are guided by gradient descent toward high-density regions $ p_t(\cdot) $ but consistently diverge from text embeddings $ c $ (triangles). This bias prioritizes probability density over semantic alignment, largely due to the reliance on CLIP~\cite{radford2021learning} as the text encoder. While aligning text with static images, CLIP struggles with the temporal dynamics of motion, hindering the diffusion model’s ability to learn a semantically coherent sampling distribution.

 A direct solution is to learn a latent space that aligns motion-text pairs and then train the diffusion model accordingly. However, the scarcity of motion-text pairs makes it difficult to train a generalized text encoder for motion, reducing the diffusion model’s generalization ability. Instead, we propose a more effective approach: leveraging an already well-aligned distribution to guide the misaligned sampling process.  Accordingly, we first estimate a reward distribution $ p_t^r(\mathbf{x}) $ from text-motion pairs, capturing semantic alignment. We then integrate this reward distribution with the vanilla sampling distribution to construct an ideal distribution $ p_t^I(\mathbf{x}) $. Crucially, our method is independent of the diffusion training process, allowing seamless integration into any diffusion model without any finetuning. As shown in Figure~\ref{fig:sampling}, sampling from this ideal distribution ensures both high-probability density and strong semantic alignment, overcoming previous limitations.

\noindent\textbf{Overview Framework.} Our framework enhances diffusion-based motion generation by constructing an ideal sampling distribution that balances motion probability with text-motion alignment. This section describes how we integrate the reward distribution into the diffusion process and sample from the resulting ideal distribution.  

Formally,  assume a reward distribution $ p_t^{r}(\mathbf{x}|c) $ has been estimated. Then we define the ideal distribution as:  
\begin{equation}
	\label{eq:p_i^t}
	p_t^{I}(\mathbf{x}|c) = p_t(\mathbf{x}|c) p_t^{r}(\mathbf{x}|c)/ Z(c),
\end{equation}
where $ Z(c) = \int p_t(\mathbf{x}|c) p_t^{r}(\mathbf{x}|c)\mathrm{d}\mathbf{x} $ is a normalizing constant. 
This formulation integrates the original sampling distribution $p_t(\mathbf{x}|c)$ with the reward distribution $p_t^{r}(\mathbf{x}|c)$, balancing both probability density and text-motion alignment.

 Using this ideal distribution, we modify the reverse process for trading-off semantic alignment and high-probability sampling as stated in the following theorem.
 \begin{theorem}\label{theorem:t1}
 When using  the ideal sampling distribution $ p_t^{I}(\mathbf{x}|c) $  in Eq. (\ref{eq:p_i^t}) to replace the vanilla sampling distribution $ p_t(\mathbf{x}|c) $, the reverse SDE becomes:  
 	\begin{equation}
 		\small
 		\begin{aligned}
 			\label{e7}
 			\mathbf{dx} \! =\! \Big[\mathbf{f}(\mathbf{x},t)- g (t)^2 \nabla \big( {\log {p_t(\mathbf{x}|c)}}  {+\log p_t^{r}(\mathbf{x}|c)} \big)\Big]\mathrm{d}t + g (t) \mathrm{d}\mathbf{w}.
 		\end{aligned}
 	\end{equation}   
 \end{theorem}  

Theorem~\ref{theorem:t1} shows that the gradient of the ideal sampling distribution decomposes into the gradients of $ p_t(\mathbf{x}|c) $ and $ p_t^{r}(\mathbf{x}|c) $. Since $ p_t(\mathbf{x}|c) $ is already known, the estimated reward distribution can directly guide the sampling process toward the ideal distribution. Next, we detail the estimation of the reward distribution and outline the motion sampling procedure.

\begin{figure}[tbp]
	\centering
	\includegraphics[width=\linewidth]{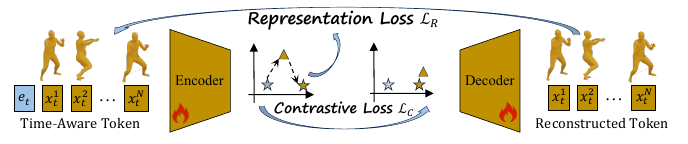}
	\caption{Framework of step-aware reward model. During this process, time-aware tokens, consisting of timestep embedding $t$ and motion embeddings $x_t^k$, are aligned with text embedding $c$ in the latent space and reconstructed via the decoder, with the encoder and decoder jointly optimized by contrastive loss $\mathcal{L}_C$ and representation loss $\mathcal{L}_R$ \cite{petrovich2022temos}.}
	\label{fig:step_aware}
     \vspace{-10pt}
\end{figure}

\subsection{Step-Aware Alignment for Reward Distribution}  
\label{sec:432}
A core challenge in estimating the reward distribution $ p_t^{r}(\mathbf{x}|c) $ is achieving precise motion-text alignment under varying noise levels in the reverse diffusion process~\cite{liang2025aesthetic, trace}. Existing methods \cite{petrovich23tmr, gmd, lamp} assume clean and noise-free motion, and overlook timestep-dependent distortions, resulting in coarse and inconsistent alignments. This misalignment hinders accurate reward estimation, which is critical for guiding sampling toward semantically faithful motion generation. To address this, we introduce a step-aware reward model for noise-adaptive alignment and a motion-to-motion reward to ensure consistency with real-world motion patterns implied by text. These components are integrated into a unified reward distribution to enhance alignment and motion quality.

\noindent\textbf{Step-Aware Reward Model.}  To mitigate timestep-dependent misalignment, we introduce a step-aware reward model $ R(\cdot)_\varphi $ illustrated in Figure \ref{fig:step_aware}, which explicitly accounts for noise variations across diffusion timesteps. Unlike conventional alignment models \cite{petrovich23tmr, lamp}, our approach incorporates a timestep token $ [e_t] $ into the motion representation, allowing the model to learn noise-dependent alignment patterns. Given an $ N $-frame motion sequence $ [x^1_t, x^2_t, \dots, x^N_t] $, we augment it with the timestep token to form the enriched representation $ [e_t, x^1_t, x^2_t, \dots, x^N_t] $. This enables the transformer-based encoder to process motion dynamics while adapting to different noise levels.

During training, noise is added to motion at timestep $ t $, and the step-aware reward model $ R_\varphi (\mathbf{x}_t,c) $ is optimized by two complementary losses: a representation loss $ \mathcal{L}_R $ \cite{petrovich2022temos} to learn meaningful motion embeddings, and a contrastive loss $ \mathcal{L}_C $ \cite{infonce} to ensure accurate motion-text retrieval. The overall training loss $\mathcal{L}_{RM}(\varphi; \mathbf{x}_t, c)$ is defined as:
\begin{equation}
	\label{eq:rm_training_loss}
	\mathcal{L}_{RM}(\varphi; \mathbf{x}_t, c) = \mathcal{L}_{C}(\varphi; \mathbf{x}_t, c) + \mathcal{L}_{R}(\varphi; \mathbf{x}_t, c).
\end{equation}  
Algorithm \ref{alg:rmtraining} detail the training procedure of the step-aware reward model.

\begin{algorithm}[t!]
\caption{Training Step-Aware Reward Model}
\label{alg:rmtraining}
\begin{algorithmic}[1]

\renewcommand{\algorithmicrequire}{\textbf{Input:}} 
\renewcommand{\algorithmicensure}{\textbf{Output:}} 
\REQUIRE 
        Step-aware reward model $R_\varphi$, training set $\mathcal{D}_{tr}$, timestep T range $[t_{\text{min}}, t_{\text{max}}]$, probability parameter $\omega$, noise scheduler $\alpha$.
\ENSURE Step-aware reward model $R_\varphi$.
\STATE $\textbf{repeat}$
\FOR{$(\mathbf{x},c)$ in $\mathcal{D}_{tr}$}
    \STATE $t \gets 0$ \Comment{Initialize $t$}
    
    \IF {$\text{Uniform(0,1)}>\omega$} 
        \STATE $t \gets \text{Uniform($t_{\text{min}}, t_{\text{max}}$)}$ \Comment{Add noise to motion}
    \ENDIF
        \STATE $\mathbf{x}_t \sim\mathcal{N}(\sqrt{\alpha}_t\mathbf{x}, (1-\alpha_t)\mathbf{I})$  \Comment{Forward process}  
        \STATE $\mathcal{L}_{RM}(\varphi; \mathbf{x}_t, c)$ by Eq. (\ref{eq:rm_training_loss}) \Comment{Compute loss}
    \STATE $\varphi \gets \varphi -  \nabla_{\varphi}\mathcal{L}_{\text{RM}}(\varphi) $ \Comment{Update parameter}
\ENDFOR
\STATE $\textbf{until}$ converged
\end{algorithmic}
\end{algorithm}

Once trained, the step-aware reward model establishes a well-aligned latent space. Given a motion $ \mathbf{x} $ and text condition $ c $, it evaluates their semantic alignment as:  
\begin{equation}
	R_{\varphi} (\mathbf{x}, c) = \cos(\mathbf{z_x}, \mathbf{z}_c),
\end{equation}   
where $ \mathbf{z_x} $ and $ \mathbf{z}_c $ are the respective motion and text embeddings  in the learned latent space.

\noindent\textbf{Motion-to-Motion Reward.} 
While text-to-motion alignment is essential, text descriptions often exhibit ambiguity, leading to inconsistencies in generated motions. To mitigate this, we introduce a motion-to-motion reward, which evaluates alignment by comparing the generated motion $ \mathbf{x}_t $ with a reference motion $ \mathbf{x}^c $ retrieved from the training set $ \mathcal{D}_{tr} $. The step-aware reward model is used to select $ \mathbf{x}^c $ as the closest match to the text condition $ c $:  
\begin{equation}
	\mathbf{x}^c = \arg \max_{\mathbf{x}\in\mathcal{D}_{tr}} R_\varphi(\mathbf{x}, c).
\end{equation}
This retrieved motion $\mathbf{x}^c$ acts as a dynamic anchor, ensuring that generated motions remain faithful to real-world motion patterns implied by the text. Accordingly, 
The motion-aligned reward is then computed as:  
\begin{equation}
	R_m\left(\mathbf{x}_t, c\right) = \cos(\mathbf{z_x}, \mathbf{z_{x^c}}),
\end{equation}  
where $ \mathbf{z_x} $ and $ \mathbf{z_{x^c}} $ are the embeddings of the generated and retrieved motions, respectively. This ensures generated motions adhere to real-world motion patterns while maintaining semantic consistency.

\noindent\textbf{Reward Distribution.} With both the step-aware reward model and the motion-to-motion reward,   we define the dual-alignment reward as:  
\begin{equation}
	R \left( \mathbf{x}_t, c\right) = \mu R_\varphi(\mathbf{x}_t, c) + \eta R_m\left(\mathbf{x}_t, c\right),
	\label{eq:reward_func}
\end{equation}  
where $ \mu $ and $ \eta $ control the contributions of text-based and motion-based alignment. This reward formulation defines the reward distribution over noised motion as:  
\begin{equation}\label{eq:estimate_pr}
	p^r_t(\mathbf{x}_t|c) = \exp \left(R \left(\mathbf{x}_t, c\right) \right) / {Z^r(c)}. 
\end{equation}  
Here, $ Z^r(c)=\int \exp(R_\varphi (\mathbf{x}, c))\mathrm{d}\mathbf{x} $ is for  normalization.

By integrating text-motion and motion-motion alignment, our approach constructs a robust reward signal that captures both semantic consistency and motion coherence. This enables more precise guidance of the diffusion sampling process, ensuring that generated motions are not only probable but also faithful to their textual descriptions.

\subsection{Reward-Guided Sampling}
\label{sec:433}

\begin{algorithm}[t!]
\caption{Reward-Guided Denoise Process}
\label{alg:t2m}
\begin{algorithmic}[1]

\renewcommand{\algorithmicrequire}{\textbf{Input:}} 
\renewcommand{\algorithmicensure}{\textbf{Output:}} 
\REQUIRE 
        Diffusion model $\epsilon_\theta$, reward model $R$, training set $\mathcal{D}_{tr}$, condition $c$, timestep $T$.
\ENSURE Generated motion $\mathbf{x}_0$. 
\STATE $\mathbf{x}_T \sim \mathcal{N}(\mathbf{0}, \mathbf{I})$ 
\STATE $\mathbf{x}^c = \arg \max_{x\in\mathcal{D}_{tr}} R_\varphi(\mathbf{x}, c)$ 
\FOR{$t = T, \cdots, 1$}
    \STATE use $\mathbf{x}^c$ to obtain reward score
    \STATE $\epsilon \sim \mathcal{N}(\mathbf{0}, \mathbf{I})$ \textbf{if} $t>1$ \textbf{else} $\epsilon = \mathbf{0}$  
    \STATE use Eq.~\eqref{safasdfad} to generate $\mathbf{x}_{t-1}$ 
\ENDFOR
\RETURN $\mathbf{x}_0$
\end{algorithmic}
\end{algorithm}

Building on the dual-alignment reward $ R(\mathbf{x}_t, c) $ and its associated distribution $ p^r_t(\mathbf{x}_t|c) $, we now integrate them into the reverse SDE to refine motion generation. The following theorem establishes how this reward distribution enhances sampling for precise text-conditioned synthesis.

\begin{theorem}\label{the:2}
Given the reward distribution $p_t^{r}(\mathbf{x}|c)$ defined in Eq. (\ref{eq:estimate_pr}), the reverse SDE can be rewritten as:
\begin{equation}
\small
\begin{aligned}
\label{t2}
    \mathbf{dx} =& \Big[\mathbf{f}(\mathbf{x},t)- g (t)^2  \nabla \big( {\log {p_t(\mathbf{x}|c)}}  +  R\left(\mathbf{x}_t, c\right) \big)\Big]\mathrm{d}t  +  g (t) \mathrm{d}\mathbf{w}.
\end{aligned}
\end{equation}
\end{theorem} 

Theorem \ref{the:2} reveals that the reward gradient $\nabla R(\mathbf{x}_t, c)$, derived from both text-aligned and motion-aligned reward components, directly influences the sampling trajectory. Integrating these gradients into the reverse SDE can dynamically steer the sampling toward a distribution that better aligns with both textual conditions and realistic structures. 

Building upon this continuous-time formulation, for practical motion generation we then derive its discrete approximation within the DDPM  \cite{ho2020denoising} framework in the following theorem.

\begin{theorem}
\label{the:t3}
Given a reverse SDE defined in Eq. (\ref{t2}), adopting standard DDPM settings \cite{songscore, ho2020denoising} where $\mathbf{f}(\mathbf{x},t) = -\frac{1}{2}\bar{\beta}_{t+\Delta t}\mathbf{x}_{t}$, $ g(t) = \sqrt{\beta_{t+\Delta t}}$, and $\bar{\beta}_t=\frac{\beta_{t+\Delta t}}{\Delta t}$, with time steps $N\rightarrow \infty$ and step size $\Delta t = \frac{1}{N}$, the reward-guided denoising process is given by:
\begin{equation}
	\small
	\begin{aligned}
		\mathbf{x}_{t- 1}= 
		\frac{1}{\sqrt{\alpha_t}} \left(  \mathbf{\bar{x}}_{t- 1} + \sqrt{\beta_t} \epsilon\right)  + \frac{\beta_t}{\sqrt{\alpha_t}} \nabla R\left(\mathbf{x}_t, \,c\right),
	\end{aligned}
\end{equation}
where $	\mathbf{\bar{x}}_{t- 1}=   \mathbf{x}_t - \frac{\beta_t}{\sqrt{1-\bar{\alpha}_t}} \epsilon_\theta(\mathbf{x}_t, t, c) $, $\beta_t$ and $\alpha_t$ are the noise schedule parameters, $\epsilon_\theta(\cdot)$ represents the diffusion model network, and $\epsilon$ is Gaussian noise sampled from $\mathcal{N}(\mathbf{0}, \mathbf{I})$.
\end{theorem}
Theorem \ref{the:t3} demonstrates that the reward gradient $\nabla R(\mathbf{x}_t, c)$, weighted by $\frac{\beta_t}{\sqrt{\alpha_t}}$, progressively influences the denoising process, adapting the sampling trajectory toward a distribution that reflects the intended motion semantics. 

To ensure the sampling stability, we remove the weight $\frac{\beta_t}{\sqrt{\alpha_t}}$ on the reward term, leading to a revised denoising process:
\begin{equation}\label{safasdfad}
	\small
	\begin{aligned}
		\mathbf{x}_{t-1} = 
		\frac{1}{\sqrt{\alpha_t}} \left(	\mathbf{\bar{x}}_{t- 1}  + \sqrt{\beta_t} \epsilon \right) + \nabla R(\mathbf{x}_t, c).
	\end{aligned}
\end{equation}

Based on the theoretical framework above, we propose Algorithm \ref{alg:t2m}, which integrates the step-aware reward model with off-the-shelf classifier-free guidance (CFG) into the diffusion-based generation process\cite{cfgho2022classifier}.

\begin{table*}[ht]
    \centering
    \setlength{\tabcolsep}{3pt} 
    \adjustbox{max width=0.92\textwidth}{ 
    \begin{tabular}{l l l l l l l}
      \toprule
     \multirow{2}{*}{Method} & \multicolumn{3}{c}{R Precision $\uparrow$} & \multirow{2}{*}{FID $\downarrow$} & \multirow{2}{*}{MM Dist $\downarrow$} & \multirow{2}{*}{Diversity $\rightarrow$} \\
      \cmidrule(lr){2-4}
       ~& Top 1 & Top 2 & Top 3 & & & \\
      \midrule
     Real       & 0.511 & 0.703 & 0.797 & 0.002 & 2.974 & 9.503 \\
      \midrule
      T2M \shortcite{Guo2022}   & 0.455$^{\pm 0.002}$ & 0.636$^{\pm 0.003}$ & 0.736$^{\pm 0.003}$ & 1.087$^{\pm 0.002}$ & 3.347$^{\pm 0.008}$ & 9.175$^{\pm 0.002}$ \\
      MDM \shortcite{Tevet2023} & 0.455$^{\pm 0.006}$ & 0.645$^{\pm 0.007}$ & 0.749$^{\pm 0.006}$ & 0.489$^{\pm 0.047}$ & 3.330$^{\pm 0.25}$ & 9.920$^{\pm 0.083}$ \\
      T2M-GPT \shortcite{Zhang_2023_T2M_GPT} & 0.492$^{\pm 0.003}$ & 0.679$^{\pm 0.002}$ & 0.775$^{\pm 0.002}$ & 0.141$^{\pm 0.005}$ & 3.121$^{\pm 0.009}$ & 9.722$^{\pm 0.082}$\\
      ReMoDiffuse \shortcite{zhang2023remodiffuse} & 0.510$^{\pm 0.005}$& 0.698$^{\pm 0.006}$& 0.795$^{\pm 0.004}$ & 0.103$^{\pm 0.004}$& 2.974$^{\pm 0.016}$&9.018$^{\pm 0.75}$      \\
      Mo.Diffuse \shortcite{Zhang2024} & 0.491$^{\pm 0.001}$ & 0.681$^{\pm 0.001}$ & 0.775$^{\pm 0.001}$ & 0.630$^{\pm 0.001}$ & 3.113$^{\pm 0.001}$ & 9.410$^{\pm 0.049}$ \\
      OMG \shortcite{liang2024omg}  & - & - & 0.784$^{\pm 0.002}$   & 0.381$^{\pm 0.008}$ & - & 9.657$^{\pm 0.085}$ \\ 
      MotionLCM \shortcite{motionlcm}  & 0.502$^{\pm 0.003}$ & 0.698$^{\pm 0.002}$ & 0.798$^{\pm 0.002}$ & 0.304$^{\pm 0.012}$ & 3.012$^{\pm 0.007}$ & 9.607$^{\pm 0.066}$ \\
      Mo.Mamba \shortcite{zhang2025motion}& 0.502$^{\pm 0.003}$ & 0.693$^{\pm 0.002}$ & 0.792$^{\pm 0.002}$ & 0.281$^{\pm 0.011}$ & 3.060$^{\pm 0.000}$  & 9.871$^{\pm 0.084}$\\ 
      CoMo \shortcite{Huang2024CoMo}& 0.502$^{\pm 0.002}$ & 0.692$^{\pm 0.007}$ & 0.790$^{\pm 0.002}$ & 0.262$^{\pm 0.004}$ & 3.032$^{\pm 0.015}$ & 9.936$^{\pm 0.066}$ \\
      ParCo \shortcite{Zou2025ParCo}  & 0.515$^{\pm 0.003}$ & 0.706$^{\pm 0.003}$ & 0.801$^{\pm 0.002}$ & 0.109$^{\pm 0.005}$ & 2.927$^{\pm 0.008}$ & 9.576$^{\pm 0.088}$ \\
      MARDM \shortcite{meng2024rethinking} & 0.500$^{\pm 0.004}$ & 0.695$^{\pm 0.003}$ & 0.795$^{\pm 0.003}$ & 0.114$^{\pm 0.007}$ & - & - \\
      MG-MotionLLM  \shortcite{wu2025mg}& 0.516$^{\pm 0.002}$ & 0.706$^{\pm 0.002}$ & 0.802$^{\pm 0.003}$ & 0.303$^{\pm 0.010}$ & 2.952$^{\pm 0.009}$ & 9.960$^{\pm 0.073}$ \\ 
      EnergyMoGen  \shortcite{zhang2025energymogen}& 0.526$^{\pm 0.003}$ & 0.718$^{\pm 0.003}$ & 0.815$^{\pm 0.002}$ & 0.176$^{\pm 0.006}$ & 2.931$^{\pm 0.007}$ &  \textbf{9.500}$^{\pm 0.091}$ \\ 
      \midrule
       MLD \shortcite{Chen2023}& 0.481$^{\pm 0.003}$ & 0.673$^{\pm 0.003}$ & 0.772$^{\pm 0.002}$ & 0.473$^{\pm 0.013}$ & 3.196$^{\pm 0.010}$ & 9.724$^{\pm 0.082}$ \\
       w/ ReAlign (Ours) & 0.567$^{\pm 0.003}$ \textbf{(+17.9\%}) & 0.759$^{\pm 0.003}$ \textbf{(+12.8\%)} & 0.848$^{\pm 0.003}$ \textbf{(+9.8\%)} & 0.195$^{\pm 0.005}$\textbf{(+58.8\%)} & 2.704$^{\pm 0.007}$ \textbf{(+15.4\%)} & 9.474$^{\pm 0.068}$ \textbf{(+86.9\%)} \\
    MLD++\shortcite{motionlcm}  & 0.548$^{\pm 0.003}$ & 0.738$^{\pm 0.003}$ & 0.829$^{\pm 0.002}$ & 0.073$^{\pm 0.003}$ & 2.810$^{\pm 0.008}$ & 9.658$^{\pm 0.089}$ \\ 
      w/ ReAlign (Ours)  & \textbf{0.572$^{\pm 0.002}$} \textbf{(+4.4\%})& \textbf{0.764$^{\pm 0.002}$} \textbf{(+3.5\%}) & \textbf{0.852$^{\pm 0.001}$} \textbf{(+2.8\%}) & \textbf{0.055$^{\pm 0.003}$} \textbf{(+24.7\%}) & \textbf{2.648$^{\pm 0.008}$} \textbf{(+5.8\%}) & {9.478$^{\pm 0.055}$}  \textbf{(+83.9\%})\\ 
      \bottomrule
\end{tabular}
}
\caption{\textbf{Comparison of text-to-motion generation performance on the HumanML3D dataset.} These metrics are evaluated by the evaluator from TM2T \cite{guo2022tm2t}. The arrows $\uparrow$, $\downarrow$, and $\rightarrow$ indicate higher, lower, and closer-to-real-motion values are better, respectively. \textbf{Bold} highlights the best results. Percentages in bracket indicate improvements over respective baselines.}
\label{tab:sota_humanmld3d}
\end{table*}

\section{Experiment}
\label{sec:exp}
\subsection{Experiment Setting}

\begin{table}[htbp]
    \centering
    \setlength{\tabcolsep}{0.5pt} 
\begin{minipage}[t]{1.04\columnwidth} 
        \centering
        \adjustbox{max width=1.04\textwidth}{
        \begin{tabular}{l@{\hspace{0pt}} C{1.5cm} C{1.5cm} C{1.5cm} C{1.5cm} C{1.5cm} C{1.6cm} C{1.55cm}}
      \toprule
     \multirow{2}{*}{Method} & \multicolumn{3}{c}{R Precision $\uparrow$} & \multirow{2}{*}{FID $\downarrow$} & \multirow{2}{*}{MM Dist $\downarrow$} & \multirow{2}{*}{Diversity $\rightarrow$} \\
    \cmidrule(lr){2-4}
     & Top 1 & Top 2 & Top 3 & & & \\
    \midrule    
    Real        & 0.424 & 0.649 & 0.779 & 0.031 & 2.788 & 11.08 \\
    \midrule    
    T2M \shortcite{Guo2022} & 0.361 & 0.559 & 0.681 & 3.022 & 2.052 & 10.72 \\
    MLD \shortcite{Chen2023} & 0.390 & 0.609 & 0.734 & 0.404 & 3.204 & 10.80 \\
    T2M-GPT \shortcite{Zhang_2023_T2M_GPT} & 0.416 & 0.627 & 0.745 & 0.514 & 3.007 & 10.86 \\	
    CoMo \shortcite{Huang2024CoMo} & 0.422 & 0.638 & 0.765 & 0.332 & 2.873 & 10.95 \\
    Mo.Mamba \shortcite{zhang2025motion} & 0.419 & 0.645 & 0.765 & 0.307 & 3.021 & 11.02 \\
    ParCo \shortcite{Zou2025ParCo} & 0.430 & 0.649 & 0.772 & 0.453 & 2.820 & 10.95 \\
    \midrule
    Mo.Diffuse \shortcite{Zhang2024} & 0.417 & 0.621 & 0.739 & 1.954 & 2.958 & \textbf{11.10} \\ 
    w/ ReAlign (Ours) & 0.419 & 0.639 & 0.764 & 0.805 & 2.801 & 10.66 \\ 
    MDM  \shortcite{Tevet2023}  & 0.403 & 0.606 & 0.731 & 0.497 & 3.096 & 10.74 \\ 
    w/ ReAlign (Ours) & \textbf{0.451} & \textbf{0.664} & \textbf{0.784} & \textbf{0.276} & \textbf{2.775} & 10.76 \\ 
      \bottomrule
\end{tabular}
}
\caption{\textbf{Comparison of text-to-motion generation performance on the KIT-ML dataset.} \textbf{Bold} highlights the best results.}
\label{tab:sota_kit}
\end{minipage}
\vspace{-10pt}
\end{table}

\noindent \textbf{Datasets and Evaluation Metrics.} We employ two widely used text-to-motion datasets, HumanML3D~\cite{Guo2022} and KIT-ML~\cite{plappert2016kit} for evaluation purposes. Consistent with the majority of prior studies~\cite{Guo2024momask, lamp}, we adopt R-Precision for Top $k$, Fréchet Inception Distance (FID), Multi-Modal Distance (MM Dist), and Diversity as evaluation metrics to assess the generation quality and alignment accuracy of our model.

\noindent \textbf{Implementation Details.} We employ the SkipTransformer \cite{Chen2023} as the foundational architecture for our step-aware reward model, consisting of a transformer encoder processing both text and motion inputs, alongside a motion decoder. Each component features 9 layers and 4 attention heads, with the latent space dimension fixed at 256. The training process incorporates a maximum timestep of 1000, a noisy motion probability of 0.5, and a negative filtering threshold of 0.9 to regulate the selection of negative samples. For model training, we adhere to the TMR framework \cite{petrovich23tmr}, employing a composite loss function expressed as a weighted combination $\mathcal{L}_{C}+\mathcal{L}_{R}$. Optimization is performed using the AdamW algorithm \cite{adamw}, configured with a learning rate of $10^{-4}$ and a batch size of 512, while other hyperparameters are consistent with those specified in the TMR~\cite{petrovich23tmr}.

\subsection{Results of Motion Generation and Retrieval}

\noindent \textbf{Text-to-Motion Generation.} As shown in Table~\ref{tab:sota_humanmld3d}, our reward-guided sampling, i.e., ReAlign, significantly enhances performance when integrated with state-of-the-art text-to-motion models. Specifically, by integrating our ReAlign, MLD++~\cite{motionlcm} achieves new SoTA results, with an R@3 of 85.2\% (+2.8\%), alongside a reduction in FID of 0.055 (+24.7\%) and an MM Dist to 2.648 (+5.8\%). Furthermore, our ReAlign also significantly enhances the performance of MDM \cite{Chen2023}, yielding SoTA results on the KIT-ML dataset, with an R@3 of 78.4\% (+7.3\%), alongside a reduction in FID of 0.276 (+44.5\%) and an MM Dist to 2.775 (+10.4\%). These consistent improvements over the baseline without ReAlign demonstrate the effectiveness of our reward-guided sampling in enhancing text-motion alignment quality. 

\begin{table*}[tb]
    \centering
    \vspace{-0.3cm}
    \renewcommand{\arraystretch}{1.1}
    \resizebox{0.9\textwidth}{!}{
    \begin{tabular}{c|lc|ccccc|ccccc}
        \toprule
        \multirow{2}{*}{} & \multirow{2}{*}{Methods} & \multirow{2}{*}{Noise} & \multicolumn{5}{c|}{Text-Motion Retrieval$\uparrow$} & \multicolumn{5}{c}{Motion-Text Retrieval$\uparrow$} \\
        & & & R@1 & R@2 & R@3 & R@5 & R@10 & R@1 & R@2 & R@3 & R@5 & R@10 \\
        \midrule
        \multirow{4}{*}{\rotatebox{90}{HumanML3D}} 
        & TEMOS \shortcite{petrovich2022temos} & \ding{55} & 40.49 & 53.52 & 61.14 & 70.96 & 84.15 & 39.96 & 53.49 & 61.79 & 72.40 & 85.89\\
        & T2M \shortcite{guo2022tm2t} & \ding{55} & 52.48 & 71.05 & 80.65 & 89.66 & \textbf{96.58} & 52.00 & 71.21 & 81.11 & 89.87 & \underline{96.78}\\
        & TMR \shortcite{petrovich23tmr} & \ding{55} & 67.16 & 81.32 & 86.81 & 91.43 & 95.36 & 67.97 & 81.20 & 86.35 & 91.70 & 95.27\\
         & LaMP \shortcite{lamp} & \ding{55} & 67.18 & 81.90 & {87.04} & \textbf{92.00} & 95.73 & {68.02} & {82.10} & {87.50} & {92.20} & \textbf{96.90}\\
        & {ReAlign (ours)} & {\ding{51}} & \textbf{67.59} & \textbf{82.24} & \textbf{87.44} & \underline{91.97} & \underline{96.28} & \textbf{68.94} & \textbf{82.86} & \textbf{87.95} & \textbf{92.44} & \underline{96.28}\\
        \midrule
        \multirow{4}{*}{\rotatebox{90}{KIT-ML}} 
        & TEMOS \shortcite{petrovich2022temos} & \ding{55} & 43.88 & 58.25 & 67.00 & 74.00 & 84.75 & 41.88 & 55.88 & 65.62 & 75.25 & 85.75\\
        & T2M \shortcite{guo2022tm2t} & \ding{55} & 42.25 & 62.62 & 75.12 & 87.50 & 96.12 & 39.75 & 62.75 & 73.62 & 86.88 & 95.88\\
        & TMR \shortcite{petrovich23tmr} & \ding{55} & 49.25 & 69.75 & 78.25 & 87.88 & 95.00 & 50.12 & 67.12 & 76.88 & 88.88 & 94.75\\
        & {ReAlign (ours)} & {\ding{51}} & \textbf{52.84} & \textbf{71.66} & \textbf{82.96} & \textbf{91.19} & \textbf{97.59} & \textbf{52.98} & \textbf{72.87} & \textbf{84.38} & \textbf{92.61} & \textbf{96.87}\\ 
        \bottomrule
    \end{tabular}
    }
    \caption{\textbf{Comparison of Text-to-motion (\textbf{left}) and motion-to-text (\textbf{right}) retrieval methods on the HumanML3D and KIT-ML datasets.} The column ``Noise''  indicates whether the method can handle noisy motion from the denoised process.}
    \label{retrieval}
    \vspace{-5pt}
\end{table*}
\noindent \textbf{Motion-Text Retrieval.} Following the small‑batch protocol of TMR\cite{petrovich23tmr}, we evaluate our ReAlign on text–motion retrieval and benchmark it against the latest state of the art. As summarized in Table~\ref{retrieval}, ours attains 67.59\% R@1 and 87.44\% R@3 on HumanML3D for the text‑to‑motion retrieval, and reaches 68.94\% R@1 and 82.86\% R@2 in the motion‑to‑text retrieval, consistently surpassing LaMP \cite{lamp} and TMR\cite{lamp}. On KIT‑ML, our approach further pushes performance to 91.19\% R@5 and 84.38\% R@3, consistently surp
assing baselines. We attribute these improvements to our noise augmentation strategy, which alleviates the limited motion diversity and text annotations in both datasets that lead to many hard negative samples, thereby enhancing the model's discriminative capability for subtle motion variations.

\begin{table}[tb]
\setlength{\tabcolsep}{0.5pt} 
  \centering
    \resizebox{0.5\textwidth}{!}{
  
  \begin{tabular}{l@{\hspace{2pt}} l l l l l l}
    \toprule
    \multirow{2}{*}{Method} & \multicolumn{3}{c}{R Precision $\uparrow$} & \multirow{2}{*}{FID $\downarrow$} & \multirow{2}{*}{MM Dist $\downarrow$} & \multirow{2}{*}{Diversity $\rightarrow$} \\
    \cmidrule(lr){2-4}& {Top 1} & {Top 2} & {Top 3}  & & & \\
    \midrule
Real & 0.511 & 0.703 & 0.797 & 0.002 & 2.974 & 9.503 \\
\midrule
MDiff \shortcite{Zhang2024} & 0.491 & 0.681 & 0.775 & 0.630 & 3.113 & 9.410 \\
w/ ReAlign & 
0.534$_{\textbf{+8.8\%}}$ & 
0.733$_{\textbf{+7.6\%}}$ & 
0.829$_{\textbf{+7.0\%}}$ & 
0.370$_{\textbf{+41.3\%}}$ & 
2.807$_{\textbf{+9.9\%}}$ & 
9.372$_{\textbf{-40.9\%}}^{\textbf{-0.04}}$ \\
\midrule
MDM \shortcite{Tevet2023} & 0.455 & 0.645 & 0.749 & 0.489 & 3.330 & 9.920 \\
w/ ReAlign & 
0.470$_{\textbf{+3.3\%}}$ & 
0.677$_{\textbf{+5.0\%}}$ & 
0.789$_{\textbf{+5.3\%}}$ & 
0.325$_{\textbf{+33.5\%}}$ & 
3.129$_{\textbf{+6.0\%}}$ & 
9.355$_{\textbf{+64.5\%}}^{\textbf{+0.27}}$ \\
\midrule
MLD \shortcite{Chen2023} & 0.481 & 0.673 & 0.772 & 0.473 & 3.196 & 9.724 \\
w/ ReAlign & 
0.567$_{\textbf{+17.9\%}}$ & 
0.759$_{\textbf{+12.8\%}}$ & 
0.848$_{\textbf{+9.8\%}}$ & 
0.195$_{\textbf{+58.8\%}}$ & 
2.704$_{\textbf{+15.4\%}}$ & 
9.474$_{\textbf{+86.9\%}}^{\textbf{+0.19}}$ \\
\midrule
MLCM$^4$ \shortcite{motionlcm} & 0.502 & 0.698 & 0.798 & 0.304 & 3.012 & 9.607 \\
w/ ReAlign & 
0.540$_{\textbf{+7.6\%}}$ & 
0.739$_{\textbf{+5.9\%}}$ & 
0.833$_{\textbf{+4.4\%}}$ & 
0.273$_{\textbf{+10.2\%}}$ & 
2.797$_{\textbf{+7.1\%}}$ & 
9.683$_{\textbf{-73.1\%}}^{\textbf{-0.08}}$ \\
\midrule
MLD++ \shortcite{motionlcm} & 0.548 & 0.738 & 0.829 & 0.073 & 2.810 & 9.658 \\
w/ ReAlign & 
0.572$_{\textbf{+4.4\%}}$ & 
0.764$_{\textbf{+3.5\%}}$ & 
0.852$_{\textbf{+2.8\%}}$ & 
0.055$_{\textbf{+24.7\%}}$ & 
2.648$_{\textbf{+5.8\%}}$ & 
9.478$_{\textbf{+83.9\%}}^{\textbf{+0.13}}$ \\
    \bottomrule
  \end{tabular}}
  \caption{\textbf{Performance improvement of motion generation methods with our step-aware reward guidance.} Results are reported on the HumanML3D dataset, showing improvements over baseline methods.}
  \label{tab:PlugAndPlay_H3D}
   \vspace{-10pt}
\end{table}

\noindent \textbf{Plug-and-Play Capability of ReAlign.}
To demonstrate the plug-and-play capability and generalizability of our ReAlign, we integrate it into various diffusion-based models for text-to-motion generation, as shown in Table~\ref{tab:PlugAndPlay_H3D}. Across methods such as Mo.Diffuse~\cite{Zhang2024}, MDM~\cite{Tevet2023}, MLD~\cite{Chen2023}, MotionLCM and its extension MLD++~\cite{motionlcm}. Our ReAlign consistently enhances performance. Notably, it achieves substantial improvements in alignment quality and motion realism, with relative gains of up to 17.9\% in R@1 and 58.8\% in FID for MLD.  
While diversity slightly decreases in some cases, this is expected and beneficial. Better diversity does not always indicate better quality, as it simply reflects motion variety. ReAlign prioritizes well-aligned motions over misaligned ones, leading to significant gains in other metrics without compromising generation quality. These results underscore the plug-and-play capability of this module, effectively elevating the efficacy of diverse motion generation frameworks.

\begin{table}[tb]
\vspace{-5pt}
  \centering

  \resizebox{0.5\textwidth}{!}{
  \begin{tabular}{c c c c c c c c c c}
    \toprule
    \multirow{2}{*}{T2M} & \multirow{2}{*}{M2M} & \multirow{2}{*}{SA} & \multicolumn{3}{c}{R Precision $\uparrow$} & \multirow{2}{*}{FID $\downarrow$} & \multirow{2}{*}{MM Dist $\downarrow$} & \multirow{2}{*}{Diversity $\rightarrow$} \\
        \cmidrule(lr){4-6}
        & & & {Top 1} & {Top 2} & {Top 3}  & & & \\
     \midrule
    \ding{55} & \ding{55} & \ding{55} 
    & 0.481 & 0.673 & 0.772 & 0.473 & 3.196 & 9.724 \\
    \midrule
    \ding{51} & \ding{55} & \ding{55} 
    & 0.556 & 0.747 & 0.841 & 0.213 & 2.761 & \textbf{9.516} \\
    \ding{55} & \ding{51} & \ding{55} 
    & 0.517 & 0.721 & 0.809 & 0.205 & 2.932 & 9.455 \\
    \ding{51} & \ding{51} & \ding{55}   
    & 0.556 & 0.750 & 0.840 & \underline{0.199} & 2.750 & 9.529 \\
    \midrule
    \ding{51} & \ding{55} & \ding{51} 
    & \textbf{0.568} & \textbf{0.761} &\textbf{0.850} & {0.212} & \underline{2.714} & 9.598 \\
    \ding{55} & \ding{51} & \ding{51}
    & 0.523 & 0.709 & 0.810 & 0.203 & 2.963 & \underline{9.525} \\
    \ding{51} & \ding{51} & \ding{51}   
    & \underline{0.567} & \underline{0.759} & \underline{0.848} & \textbf{0.195} & \textbf{2.704} & 9.474\\
    \bottomrule
    \end{tabular}
    }
    \vspace{-5pt}
      \caption{\textbf{Ablation study of the text-to-motion on HumanML3D dataset.} ``T2M", ``M2M" and ``SA" denote the text-to-motion reward, motion-to-motion reward and step-aware training, respectively.}
      \label{tab:ablation_reward}
\end{table}
\subsection{Ablation Studies and Discussions}

\begin{figure}[tb]
    \centering
    \includegraphics[width=1\linewidth]{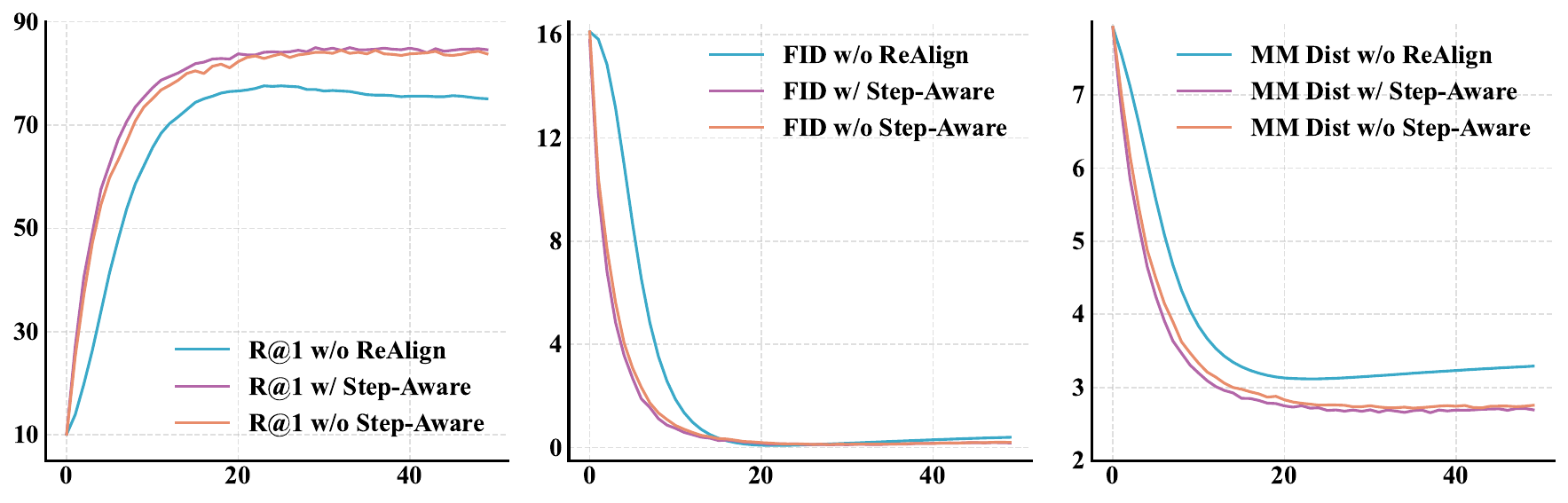}
    \caption{Comparison of motion generation quality across denoising steps for the MLD w/o ReAlign, MLD w/o Step-Aware, and MLD w/ Step-Aware (ReAlign). ReAlign consistently outperforms the others, highlighting the necessity of explicit noise handling during denoising.}
    \label{fig:handling}
\end{figure}

\noindent \textbf{Effectiveness of Handling Noisy.} To verify the necessity of handling noise and to avoid reward hacking, we varied the denoising steps of MLD from 1 to 50, employing both the reward model (RM) and the ReAlign to perform reward-guided sampling. This comples the model to generate noisy motions. As shown in Figure~\ref{fig:handling}, compared to the baseline and the RM, ours achieves superior performance across all denoising steps, demonstrating that explicitly handling noise during the denoising benefits the generation of higher quality motions. Notably, we observed that MLD generates motions with discernible semantics even in the early steps of denoising, rather than purely noise. This behavior may be attributed to the latent VAE used in MLD, which inherently exhibits robustness to noise. These experimental results demonstrates the feasibility and necessity of handling noise in denoising, aligning with conclusions drawn in DALLE-2~\shortcite{dalle2} and GLIDE~\shortcite{glide}.

\noindent \textbf{Effectiveness of Reward Sampling.} We assess the ReAlign, including T2M and M2M alignment rewards, along with the step-aware strategy in text-to-motion generation on MLD~\cite{Chen2023}. As shown in Table~\ref{tab:ablation_reward}, results indicate that the T2M reward significantly improves the alignment between the generated motions and text descriptions, as well as the realism of the motions. While the M2M reward alone exhibits limited efficacy due to the inaccuracy of text-to-motion retrieval, its integration with the step-aware strategy further enhances motion realism, validating the necessity of handling noise during the sampling. The combination of T2M and step-aware strategies achieves optimal performance, with M2M providing additional realism gains.

\begin{table}[tb]
  \centering
  \resizebox{1\linewidth}{!}{
  \begin{tabular}{cccccccc}
    \toprule
    \multirow{2}{*}{CFG} & \multirow{2}{*}{ReAlign} & \multicolumn{3}{c}{R Precision $\uparrow$} & \multirow{2}{*}{FID $\downarrow$} & \multirow{2}{*}{\makecell{MM Dist \\$\downarrow$}} & \multirow{2}{*}{\makecell{Diversity \\$\rightarrow$}} \\
    \cmidrule(lr){3-5}
    & & Top 1 & Top 2 & Top 3 & & & \\
    \midrule
    \ding{55} & \ding{55} & 0.263 & 0.407 & 0.506 & 0.586 & 4.823 & 8.613 \\
    \ding{51} & \ding{55} & 0.481 & 0.673 & 0.772 & 0.473 & 3.196 & 9.724 \\
    \ding{51} & \ding{51} & \textbf{0.567} & \textbf{0.759} & \textbf{0.848} & \textbf{0.195} & \textbf{2.704} & \textbf{9.474} \\
    \bottomrule
    \end{tabular}
    }
      \caption{\textbf{Ablation study of the guidance strategy}. Evaluation conducted on the HumanML3D with MLD~\shortcite{Chen2023} as the baseline. ``CFG" and ``ReAlign" denote the classifier-free guidance and our ReAlign, respectively.}
     \label{tab:cmp_cfg}
 \vspace{-10pt}
\end{table}

\noindent \textbf{Discussion on ReAlign and Classifier-free Guidance.} As shown in Table \ref{tab:cmp_cfg}, our ReAlign is compatible with Classifier-Free Guidance (CFG)~\cite{cfgho2022classifier}, and their integration can further unleash the performance of the diffusion model. Unlike CFG, which requires training, our ReAlign is plug-and-play and supports flexible reward function design tailored to different tasks (e.g., physical reward, trajectory reward, style reward, etc.). In this work, we primarily focus on improving text-motion alignment, while future research will explore reward designs for more tasks.

\section{Conclusion}\label{conclusion} We propose ReAlign, a plug-and-play reward-guided sampling strategy for diffusion-based text-to-motion generation. By jointly optimizing text-aligned and motion-aligned rewards during denoising, ReAlign effectively improves semantic consistency and motion realism. Our method integrates seamlessly with existing diffusion models without extra fine-tuning. Extensive experiments demonstrate that ReAlign achieves significant gains in both text-motion alignment and motion quality over state-of-the-art baselines.

\section{Acknowledgements}
This work is supported by National Natural Science Foundation of China (62302093, 62276134, 52441503), Jiangsu Province Natural Science Fund (BK20230833), Double First-Class Construction Foundation of China (23GH020227), the Fundamental Research Funds for the Central Universities (2242025K30024), the Open Research Fund of the State Key Laboratory of Multimodal Artificial Intelligence Systems (E5SP060116), and the Singapore Ministry of Education (MOE) Academic Research Fund (AcRF) Tier 1 grant (Proposal ID: 23-SIS-SMU-070). 
We thank the Big Data Computing Center of Southeast University for providing the facility support on the numerical calculations.
Any opinions, findings and conclusions or recommendations expressed in this material are those of the author(s) and do not reflect the views of the Ministry of Education, Singapore.

\bibliography{aaai2026}

\end{document}